# Heuristic Framework for Multi-Scale Testing of the Multi-Manifold Hypothesis

F. Patricia Medina, Linda Ness and Melanie Weber, Karamatou Yacoubou Djima

**Abstract** When analyzing empirical data, we often find that global linear models overestimate the number of parameters required. In such cases, we may ask whether the data lies on or near a manifold or a set of manifolds (a so-called multi-manifold) of lower dimension than the ambient space. This question can be phrased as a (multi-) manifold hypothesis. The identification of such intrinsic multiscale features is a cornerstone of data analysis and representation, and has given rise to a large body of work on manifold learning. In this work, we review key results on multi-scale data analysis and intrinsic dimension followed by the introduction of a heuristic, multiscale framework for testing the multi-manifold hypothesis. Our method implements a hypothesis test on a set of spline-interpolated manifolds constructed from variance-based intrinsic dimensions. The workflow is suitable for empirical data analysis as we demonstrate on two use cases.

## 1 Introduction

In many empirical data sets, the dimension of the ambient space exceeds the number of parameters required to parametrize local models. Geometrically, this is evident in data sets sampled from a manifold of lower dimension than the ambient space. The simplest hypothesis for explaining this observation is that the number of local

F. Patricia Medina
Worcester Polytechnic Institute; e-mail: `fpmedina@wpi.edu`

Linda Ness
Rutgers University; e-mail: `nesslinda@gmail.com`

Melanie Weber
Princeton University; e-mail: `mw25@math.princeton.edu`

Karamatou Yacoubou Djima
Amherst College; e-mail: `kyacouboudjima@amherst.edu`

parameters required to model the data is constant. We can formalize this by asking whether the data lies on or near a *d*-dimensional manifold or whether the data was sampled from a distribution supported on a manifold. This manifold hypothesis is central to the field of manifold learning. In the present article, we outline a heuristic framework for a hypothesis test suitable for computation and empirical data analysis. We consider sets of manifolds (multi-manifolds) instead of single manifolds, since empirical data is more likely to lie near a multi-manifold than on a single manifold (see, e.g., [1] ). For this, consider the following motivating question: *Given a data set in $\mathbb{R}^n$, is it on or near a multi-manifold?* Note, that the manifolds do not need to be linear; they may have different intrinsic dimensions and they may intersect.

**Proposition** (Multi-manifold Hypothesis Test)
Given a data set $X = \{x_i\}_{i \in I}$ in $\mathbb{R}^D$ and a multi-manifold $\mathscr{V}$, is the expected distance of the points in $X$ to $\mathscr{V}$ more than one would expect? If so, reject $\mathscr{V}$ as being a multi-manifold that fits $X$.

This hypothesis is closely related to the identification of intrinsic dimensions. A large body of work has been devoted to the study and computation of intrinsic dimension. If the data set can be partitioned into subsets, each of which has a single intrinsic dimension, hypothesis testing methods might be applied to the corresponding subsets separately.

## 1.1 Contributions

In the present paper, we propose a heuristic framework for testing a multi-manifold hypothesis on real-world data sets. Our method partitions a data set into subsets based on intrinsic dimension and constructs a multi-manifold whose dimensional components fit the partitions. Finally, we compute test statistics to evaluate the goodness-of-fit of a candidate multi-manifold with the data set. To our knowledge, this is the first implementable heuristic for multi-manifold hypothesis testing.

For efficiently computing intrinsic dimensions, we introduce a multi-scale variance-based notion of intrinsic dimension, denoted as $d_{VLID}$[1]. We demonstrate our method on two low-dimensional densely sampled data sets with visible geometry: One data set is a sample from a sphere-line configuration (see Fig. 2), the other a subset of a 3-dimensional image of the Golden Gate Bridge recorded with LiDAR technology. The computational experiments demonstrate that multi-scale techniques can be used to overcome the issue of linear models overestimating the dimension of the underlying data. The decomposition of the data set into subsets with a single local intrinsic dimension promises to improve the understanding of the data and provide features that could be used as preprocessed input to further analysis and machine learning tools.

---

[1] We define $d_{VLID}$ to be a pointwise statistic that depends on a set of local neighborhoods at each point. The intrinsic dimension $d$ is computed for sets of data points in each local neighborhood. Then $d_{VLID}$ is the minimum of these intrinsic dimensions. Hence points sampled from a local manifold of dimension $d$ have $d_{VLID}$ equal to $d$. A more formal definition is in Section 2.4.

Our method provides a practical heuristic for testing a manifold hypothesis as it is central to manifold learning. The introduced framework is general and can be implemented using a variety of computational tools in different parts of the workflow. Two fundamental types of statistical reasoning, hypothesis testing, and variance-based analysis, are used in combination with multiscale representation methods.

## *1.2 Outline*

We start with an extensive review of (multi-scale) techniques for dimensionality analysis and manifold learning (section 2). In section 3 we propose a heuristic, multiscale framework for testing a multi-manifold hypothesis. Section 4 describes our implementation of the framework, including a variance-based notion of intrinsic dimension that we developed as part of the workflow. We demonstrate our method on (i) a simple sphere-line configuration and (ii) imaging data obtained with LiDAR technology. The paper concludes with a list of open questions and directions that we suggest for future work.

## 2 Related Work

In this section, we review related work on manifold learning and geometric data analysis that underlie or motivate the ideas outlined in the paper.

## *2.1 Manifold Learning*

A real world data set $X$ is typically a set of $m$ vectors $x_i$, with $D$ components. Hence the data set $X$ is a subset of $m$ points in a $D$-dimensional Euclidean space, denoted $X \subset \mathbb{R}^D$. A central question in manifold learning is: Is $X$ on or near a manifold of dimension $d < D$? If so, i.e., if the manifold hypothesis is true, then it is reasonable to expect that $X$ has another representation as a subset of a space of dimension $d < D$, where $d$ may be much smaller than $D$, denoted $d \ll D$. Furthermore, because the data points are on a manifold which may be curved in its embedding space, the most natural or informative dimension reduction may be non-linear.

Many results in manifold learning are focused on dimension reduction mappings $f : X \to \mathbb{R}^d$, defined by non-linear functions of the $D$-dimensional coordinates of points in $X$. Conveniently, the mappings can be defined for general data sets X; they do not require the manifold hypothesis to be validated first. Some of the first papers that discussed this and presented examples and non-linear methods for dimension reduction are [19, 23, 43, 44, 50]. These methods could be used to infer non-linear parameters, e.g., the pose variables and azimuth lighting angle sufficient to parametrize a set of images, which would have been invisible to traditional dimension reduction techniques such as Principal Component Analysis (PCA) and Multidimensional Scaling (MDS) [50].

Laplacian Eigenmaps (LE) is a well-known example of non-linear dimension reduction mapping defined by the first few eigenfunctions of the normalized Laplacian matrix associated to a given data set. LE is used in numerous applications and is very popular in spectral clustering [40]. In [5, 7], Belkin and Niyogi justified the the LE-algorithm by proving that, when a sufficiently large data set $X$ is uniformly sampled from a low dimensional manifold $M$ of $\mathbb{R}^D$, the first few eigenvectors of the normalized Laplacian matrix $M$ are discrete approximations of the eigenfunctions of the Laplace-Beltrami operator on the manifold. Recall that the normalized Laplacian matrix $M = D^{-1}L$ where $L$ is the similarity kernel matrix whose entries $L_{i,j}$ are defined by

$$L_{i,j} = \exp\left(-\frac{||x_i - x_j||^2}{2\varepsilon}\right) \qquad (1)$$

and $D$ is the diagonal normalization matrix with entries $D_{i,i} = \sum_j L_{i,j}$.

Subsequent research has focused on non-linear dimension reductions mappings that approximately preserve distances. Using a symmetric matrix adjoint to the normalized Laplacian, Nadler, Lafon, Coifman and Kevredkides in [38] defined a non-linear dimension reduction mapping known as Diffusion Maps which approximately preserves diffusion distances. The normalized Laplacian and its symmetric adjoint are stochastic matrices and hence define random walks; the diffusion distance at time $t$ between two points $x_i$ and $x_j$ is the probability that the random walks starting at $x_i$ and $x_j$ will reach the same point at time $t$. This distance is a more accurate and robust model for the distance traveled by moving to nearby points, i.e., the distance by moving along the manifold best-fitting the data points. Diffusion maps have been applied to many types of data set, for example, in characterizing the properties of molecular dynamics [42] [52] [17] [55]. In further related developments, Rohrdanz, Zheng, Maggioni and Clementi [54] used locally scaled diffusion maps to more accurately determine reaction coordinates. Joncas, Meila and McQueen [29] developed methods for defining and computing non-linear dimension reduction mappings that approximately preserve the original metric induced by the ambient Euclidean metric. Approximate preservation of this metric would enable preservation of shape properties involving curvature. McQueen, Meila, VanderPlas and Zhang have developed and documented Megaman, a scalable publicly available software package for manifold learning from data [37]. Our intrinsic dimension algorithms demonstrate automated methods for decomposing data sets into subsets each of which lie on or near a not necessarily linear submanifold of a single dimension.

## 2.2 The (Multi-)Manifold Hypothesis

The manifold hypothesis is central to the area of manifold learning. Recent work by Fefferman, Mitter, and Narayanan [24] formulate and prove a manifold hypothesis test, thereby providing a theoretical framework for testing whether a given data set lies on or near a manifold. Narayanan and Mitter obtained bounds on the sample complexity of Empirical Risk Minimization for a class of manifolds with bounds on dimension, volume, and curvature [39].

When data is sampled from a single manifold of dimension $d$ in $\mathbb{R}^D$ with a restricted noise model, Chen, Little and Maggioni [15] have introduced Geometric Multi-Resolution Analysis (GMRA). Using a notion of geometric wavelets they show that one can construct a linear multi-manifold that gives a good local approximation to this manifold on certain scales. The local linear multi-manifold can be obtained by projecting onto the local linear subspace determined by the intrinsic dimension. GMRA exploits a dyadic tree to decompose the manifold and sampled data into pieces at each scale. The current implementation of our method also uses a dyadic tree and computes local linear approximations to the data.

Lerman and collaborators noted that empirical data is more likely to fit a set of manifolds rather than a single manifold, hence motivating the notion of multi-manifolds that we adopt here. We review recent work on multi-manifolds that motivated our approach: Arias-Castro, Chen and Lerman [1] point out that when a data set lies on or near multiple manifolds in Euclidean space, the "foremost" problem is clustering the data into subsets associated with different manifolds. They propose a Higher Order Spectral Clustering algorithm (HOSC) that applies spectral clustering to a pairwise affinity function. The algorithm provably outperforms other clustering methods (e.g., Ng, Jordan and Weiss [40]) in its accuracy on small scales and under low sampling rates. It utilizes the notion of tubular neighborhoods around manifolds and leverages the definition of correlation intrinsic dimension [26, 34] to determine the radii of these neighborhoods. The approach assumes that the data lies almost completely in these neighborhoods with the exception of a set of outliers which satisfy particular sampling assumptions. While we adopt some of these ideas, our heuristic approach does not make this assumption, nor does it assume a particular sample distribution. Additional context on multi-manifolds can be found in [53] and the references therein.

## 2.3 Quantitative Rectifiability

A challenging problem is to determine if a set of data is a subset of "nice" manifolds, i.e., is piece-wise smooth. One way to make this precise is the notion of rectifiability:

**Definition 1 (Rectifiability).** A subset $X \subset \mathbb{R}^D$ with Hausdorff dimension $d \in \mathbb{Z}$ is said to be *rectifiable* if it is contained in the union of a countable family of $d$-dimensional Lipschitz graphs with the exception of a set of Hausdorff measure zero.

A stronger quantitative condition implying rectifiability, the BPLG property, was established by David and Semmes [22]. Prior to this, Jones [30] proved a necessary and sufficient condition for a subset of the plane to be contained in a plane curve (i.e., in the image of the unit interval under a Lipschitz mapping). He defined $\beta$-numbers for each scale and location which measure the deviation of a set from the best fitting line. He proved that the length of the curve is bounded in terms of the sum of the $\beta$-numbers. Recently, Azzam and Schul [2] proved a variant of Jones theorem for a more general case, providing bounds on the Hausdorff measure for integer-dimensional subsets of Euclidean spaces using a generalization of Jones' $\beta$-numbers.

In this paper, we did not attempt to determine if there are conditions on subsets of Euclidean spaces with a specified variance-based intrinsic dimension which would guarantee quantitative rectifiability. Jones' multi-scale techniques and statistics associated with each location and scale inspired our multi-scale definition of variance-based dimension for each locality. The approach in this paper enlarges the class of Multi-scale SVD (MSVD) unsupervised learning techniques (sometimes referred to as MLPCA) used previously to automatically generate features for supervised machine learning [3, 4, 8].

## 2.4 Stratified Space Construction

We now review the notion of stratified spaces which is used synonymously for multi-manifolds:

**Definition 2 (Stratified Space).** A stratified space is a topological space that can be decomposed into manifolds.

While the two notions are closely related, the emphasis of stratified spaces is topological. Bendich, Gasparovic, Tralie and Harer [9] used stratified spaces to develop a heuristic approach for partitioning the space. Their approach is both similar and complementary to the partitioning approach used in our methodology. It exploits previous ideas in [3, 4, 8] on multi-scale data analysis. One similarity is the use of a tree-based approach that decomposes data sets using tree structures. While they construct the tree-based decomposition using the CoverTree algorithm [10] with gap-based local intrinsic dimensions, we compute a fixed dyadic tree structure using variance-based intrinsic dimensions. A second similarity arises in the construction of multi-manifolds: While we focus on fitting piece-wise linear manifolds to the data on which to compute the test statistics, they summarize the decomposition into a graph structure that captures the local topology. The results of both approaches are to some extend complementary: Our fixed dyadic tree structure gives coarse-grained information on the topology of the multi-manifold. Their approach provides more refined information by exploiting persistent homology statistics to refine the stopping condition and to coalesce some of the sets in the original decomposition.

## *2.5 Intrinsic Dimension*

The problem of estimating the intrinsic dimension (ID) of a data set is a recurring topic in the analysis of large data sets that require efficient representation, i.e., representation that simplifies visualization, decreases storage needs, improves computational complexity, etc. An essential step in this problem is to uncover the true or intrinsic dimensionality of the data. Indeed, although the data may be embedded in $\mathbb{R}^D$, its intrinsic dimension, or as Fukunaga defines it [25], the minimum number $d$ such that the data set lies entirely within an $d$-dimensional subspace of $\mathbb{R}^D$, is often much smaller than $D$. From this point of view, intrinsic dimensionality estimation can be put under the general umbrella of dimension reduction.

The intrinsic dimension of a data set can be estimated globally and locally. Global estimation methods assume that there is only one dimension for the entire data set. By contrast, local estimation methods assume that the dimension differs from one region of the data set to another and, therefore, the dimension is computed for each data point in relation to its neighbors.

In our work, we focus on local estimation of intrinsic dimensionality; however, it is important to note that several local techniques are obtained by adapting a global technique to small regions or points in a large data set. We will often use intrinsic dimension of a point to refer to the local intrinsic dimension of the data set centered at the said point. This abuse of language is common in dimensionality estimation; points are not regarded as zero-dimensional objects but rather as carrying the dimensionality of a region large enough to accurately capture the surrounding manifold but small enough to preserve a notion of locality.

In the following, we review a few important estimation techniques:

**Projection-based Methods.** The goal of projection-based methods is to find the best subspace $\mathbb{R}^d$ on which to project a data set embedded in $\mathbb{R}^D$. The criteria for best subspace is often encoded by an error or cost function that one seeks to minimize. For example, PCA, a very popular linear projection technique, minimizes the reconstruction error between a data matrix and its reconstruction, which is the projection onto basis vectors that represent the directions of greatest variance of the data. The PCA algorithm for estimating intrinsic dimension is as follows:

1. Compute the eigenvalues $\lambda_1 \ldots, \lambda_D$ of the $D \times D$ data covariance matrix and order them from highest to lowest.
2. Compute the (percent) cumulative sum of the first $k$ eigenvalues $100 \left( \sum_{i=1}^{k} \lambda_i \right) / \left( \sum_{i=1}^{D} \lambda_i \right)$. These cumulative sums are fractions of the total variance explained by the corresponding eigenvalues.
3. Define the intrinsic dimension $d$ as the number of non-null eigenvalues whose cumulative sum is larger than a prescribed threshold value, e.g., 95%.

Even though PCA remains a go-to technique in dimensionality reduction, it has several known issues such as its lack of robustness to noise or its overestimation of the intrinsic dimension in global settings for certain data sets, in particular, those that are non-linear. For instance, PCA characterizes the $d$-dimensional sphere as being $d+1$-dimensional. To resolve this issue, several non-linear techniques such as neural networks or many other methods from the manifold-recovering toolbox, e.g., Kernel PCA [45], Laplacian Eigenmaps [6], Diffusion Maps [18], have been developed. More information about these techniques, which are sometimes called kernel methods, can be found in [27, 33].

In 1971, Fukunaga and Olsen developed a local intrinsic dimension estimation method based on PCA. To achieve this, they create Voronoi sets in the data sets using a clustering algorithm and compute each set's intrinsic dimension using the algorithm described earlier. There are many improvements on this local PCA, including the MSVD method by Little and Maggioni which we describe next.

**Multiscale Methods.** Another method based on singular value decomposition is the Multiscale Singular Value Decomposition (MSVD) method of Anna Little and Mauro Maggioni. MSVD is a multiscale approach to determining intrinsic dimension, but it can also be classified as a projection method. In particular, the main difference between this method and the local PCA of Fukunaga and Olsen is that the local PCA algorithm computes the intrinsic dimension using a fixed scale determined interactively, while MSVD estimates the intrinsic dimension by studying the growth the Squared Singular Values (SSV's) in function of changes in scale [15, 35]. MSVD is based on the observation that for small scales $r$, SSV's representing the tangential space, i.e., the intrinsic dimension have a linear relationship with $r$, while SSV's representing the curvature space have a quadratic relationship with $r$. For large scales, SSV's representing the tangential space have a quadratic linear relationship with $r$, while SSV's representing the curvature space have a quartic relationship with $r$.
In absence of noise, the MSVD algorithm can be summarized as follows: given a data set $X = \{x_1, \ldots x_N\} \subseteq \mathbb{R}^D$ and a range of scales or radii $r_1, \ldots, r_p$,

(i)   construct a ball $B_{r_j}(x_i)$ of radius $r_j$ centered at $x_i$, $i = 1, \ldots, N$, $j = 1, \ldots, p$.
(ii)  compute the SSV $\lambda_k^2(x_i, r_j)$, $k = 1, \ldots, D$ for each ball $B_{r_j}(x_i)$.
(iii) for each point $x_j$, use a least-square regression of $\lambda_k^2$ as a function of $r$ to discriminate the curvature tangential from the tangential ones.
(iv)  the intrinsic dimension $d$ is defined as the number of tangential SSV's.

In presence of noise, an extra step is added to eliminate certain values of $r$ where the noise creates variability in SSV's that can not be attributed to dimensionality. In [15], the authors implemented the MSVD algorithm on both artificial manifolds and real world data sets and obtained excellent results.

**Fractal-Based Methods.** These techniques estimate the intrinsic dimension based on the box-counting dimension, which is itself a simplified version of the Hausdorff dimension. Consider the data set $X \subseteq \mathbb{R}^D$ and let $v(r)$ be the minimal number of boxes of size $r$ needed to cover $X$. The underline{box counting dimension} $d$ of $X$ is defined as

$$d := \lim_{r \to 0} \frac{\ln(v(r))}{\ln(1/r)}. \qquad (2)$$

The box-counting dimension estimation is computationally prohibitive, therefore many methods such as the correlation dimension attempt to give a computationally feasible approximation. This dimension estimate is based on the correlation integral:

$$C(r) := \lim_{N \to \infty} \frac{2}{N(N-1)} \sum_{i=1}^{N} \sum_{j=i+1}^{N} \mathbb{1}_{\{\|x_j - x_i\| \leq r\}}, \qquad (3)$$

where $x_1, \ldots, x_N$ are $N$ i.i.d. samples which lie on $X$. Given $C(r)$, the corresponding correlation dimension is given by

$$d_C \approx \lim_{r \to 0} \frac{\ln(C(r))}{\ln(r)}. \qquad (4)$$

The GP algorithm, named after its creators, Grassbered and Procaccia, estimates $d$ by finding the slope of the linear part of the plot of $\ln(C(r))$ versus $\ln(r)$. This decreases the sensitivity of the algorithm to the choice of $r$. However, their method is still computationally expensive as one needs $N > 10^{d_C/2}$ data points to obtain an accurate estimate of the intrinsic dimensionality. In 2002, Camastra and Vinciarelli proposed a fractal adaptation of the GP method that could be use for smaller data sets $X$ [13]. The algorithm starts by generating data sets $Y_i$, $i = 1, \ldots m$ of the same size as $X$ for which the intrinsic dimensionality $d_i$ is known. Using the GP method, they compute the correlation dimension $d_C^{(i)}$ of each data set and create a reference curve which is the best fitting curve to the data set $\{(d_i, d_C^{(i)}) : i = 1, \ldots m\}$. Then, they determine the correlation dimension $d_C$ for $X$ and using the reference curve, find the corresponding intrinsic dimension $d$. This heuristic method is based on the assumption that the reference curve depends on $N$ but is not affected by the type of data set $Y_i$ used in its construction.

Several other fractal-based methods were also developed to improve GP. The method of surrogate data consists in computing the correlation dimension for a (surrogate) data set with size larger than $X$ but with the same statistical properties (mean, variance and Fourier Spectrum), in the spirit of the bootstrap method [51]. Takens's method improves the expected error in the GP algorithm and is based on Fisher's method of Maximum Likelihood [49].

Other estimators of intrinsic dimension based on the correlation integral are based on applying the maximum likelihood estimation (MLE) principle to the distances between data points. In their 2005 paper, Levina and Bickel assume that the observations within a specified radius of $x$ are sampled from Poisson process and estimate the intrinsic dimension of the Poisson process approximation via some statistical measures [34]. Other MLE based-methods include an extension of Levina's and Bickel's work in [28], where the authors model the data set as a process of translated Poisson mixtures with regularizing restrictions in the presence of noise.

**Nearest Neighbor-based Methods.** Suppose we are given data points $X = \{x_1, \ldots, x_N\} \subset \mathbb{R}^D$ drawn according to an unknown density $p(x)$. Assume that this subset $X$ is of intrinsic dimension $d$. Let $V_d$ be the volume of the unit sphere in $\mathbb{R}^D$ and denote by $R_k(x)$ the distance a point $x$ and its $k^{th}$ nearest neighbor. The intrinsic density of $X$ can be approximated by the formula [34]:

$$\frac{k}{N} \approx p(x) V_d R_k(x)^d.$$

Based on this formula, Pettis and al. [41] show that, with some additional assumptions, the intrinsic dimensionality $d$ and $k$ are related by

$$\frac{1}{d} \ln k \approx \ln \left( \mathbb{E}[\overline{R}_k] \right) + C, \tag{5}$$

where $C$ is a constant and $\overline{R}_k$ is evaluated using

$$\overline{R}_k = \frac{1}{N} \sum_{i=1}^{N} R_k(x_i).$$

Then, one uses linear regression to plot $\ln k$ versus $\ln \left( \mathbb{E}[\overline{R}_k] \right)$ and the $d$ is estimated as the reciprocal of the slope of this line.

Another method based on nearest-neighbors is the Geodesic Minimal spanning tree (GMST), which estimates the intrinsic dimensions by 1) finding the geodesic distances between all points in a data sets, 2) constructing a similarity matrix based on these distances and 3) computing a minimal spanning subgraph from which the intrinsic dimension is estimated [21]. A major drawback of Nearest-Neighbors-based approaches is their large negative bias due to under sampling. Improvements were obtained by giving more weights to interior points and forcing constant dimension in small neighborhood for local estimation [14, 20].

**Analytic Methods based on metric spaces.** Nearest neighbor search is a fundamental area in which it is also essential to estimate the intrinsic dimension. In [10], the authors mention two quantities that can be used as proxies for intrinsic dimension. The first was developed by Karger and Ruhl for classes of metrics with a growth bound [31]. Let $B_r(x)$ represent the ball of radius $r > 0$ centered at $x$. For a data set $X$, Karger and Ruhl define the expansion constant $c$ of as the smallest value $c \geq 2$ such that, for any point $x \in X$:

$$|B_r(x)| \leq c |B_{r/2}(x)|.$$

From this, assuming that $X$ is sampled uniformly on some surface of dimension $d$ which would imply $c \sim 2^d$, they define the expansion dimension $d_{KR}$ by

$$d_{KR} = \ln c.$$

However, in practice, this formula often overestimates the intrinsic dimension. For example, KR-dimension may grow arbitrarily large if one adds just a single point to a data set embedded in a Euclidean space. Another intrinsic dimension estimate comes from Krauthgamer and Lee [32] and is based on the doubling dimension or doubling constant, i.e., the minimum value $c$ such that every ball of a given radius in a set $X$ can be covered by $c$ balls of half the radius. Given $c$, the dimension $d_{KL}$ is defined as before as

$$d_{KL} = \ln c.$$

The dimension estimate $d_{KL}$ is more robust to changes in data sets than $d_{KR}$, however there are few convergence results for the algorithm [10]. When representing a dimensional clustering hierarchy (as used in the cover tree algorithm [10]), $c$ can be used to bound the number of children in the next tree level (upper bounded by $c^4$). Its value is computed by considering balls $B_1$ and $B_2$ of radius $r$ and $2r$ around each data point and counting the number of data points in each ball. Then $c$ is the smallest value, such that $|B_2| \leq c|B_1|$. Such a tree structure allows for performing a fast nearest neighbor search, $O(c^{12} \log(|X|))$, after one-time construction cost of $O(c^6|X|\log(|X|))$ and storage $O(|X|)$ [10]. Interestingly, the doubling dimension allows for a rigorous estimation of these complexity results; an approach that could be extended to the methods described below.

There are several other ideas for estimating intrinsic dimension including Multidimensional Scaling Methods (MDS), Topology representing network (TRN), Bayesian estimators and many more. A lengthier account of those presented here can be found in [35]. Camastra's survey of data dimensionality estimation gives a very good description and classification of different estimators [12]. A thorough survey of nonlinear dimensionality reduction techniques can be found in [33].

The present paper defines a variance-based notion of intrinsic dimension $d_{VLID}$, similar to the MSVD method in its multi-scale approach. Moreover, it is similar to PCA in that it exploits the principal values accounting for a prescribed proportion of the total variance (see Sec. 4.1).

## 3 Methodology

We now present a computational methodology for testing the multi-manifold hypothesis (Prop. 1). Our approach is based on a training-testing routine that constructs candidate manifolds based on one part of the data (training set) and evaluates the hypothesis through a testing procedure on the remaining data points (testing set). The workflow consists of three major steps, (i) the sampling of training and testing sets, (ii) the construction of candidate manifolds and (iii) goodness of fit statistics for evaluation.

For the first step, we either separate the data points into two groups (training/ testing) or sub-sample two sets of data points if the given data set is very large. The sampling should preserve the intrinsic geometry of the original data set, since we want to test if we can construct a candidate manifold that represents the whole data set reasonably well. To construct candidate manifolds we draw on the extensive literature on manifold learning and dimensionality analysis as detailed below.

A key step in the methodology is the evaluation of the candidate manifolds that represent the actual hypothesis test. For this, we want to estimate an approximate square distance, that is, compute shortest distances from each sample point to the candidate manifold. Formally, we evaluate the empirical loss against the loss function

$$\mathscr{L}(\mathscr{V},P) = \int d(x,\mathscr{V})^2 dP(x) \qquad (1)$$

where P is the probability distribution from which the data set is sampled. By analyzing the distribution of their deviation, i.e.

$$P\left[\sup_k |\frac{1}{|X|}\sum_{x_i \in X} d(x_i,\mathscr{V})^2 - \mathscr{L}(\mathscr{V},P)| < \varepsilon\right] > 1-\delta. \qquad (2)$$

Here, $\delta$ is the significance level (e.g., the commonly used $\delta = 0.05$) and $k$ a resolution parameter in the construction of the candidate manifold $\mathscr{V}$. However, since we cannot directly compute the loss function $\mathscr{L}$, the test statistic (eq. 2) is not suitable for computational purposes. Instead, we use the following heuristics:

$$\sup_k \frac{1}{|X|}\sum_{x_i \in X} d(x_i,\mathscr{V})^2 < \hat{\delta}, \qquad (3)$$

where $k$ is again a resolution parameter and $\hat{\delta} := \hat{\delta}(|X|,k)$ the square-distance threshold for which we are willing to accept the candidate manifold. The threshold depends on both the sample size $|X|$ and the resolution parameter $k$.

These ideas are implemented by the following workflow, shown schematically in Fig. 1:

Step 1 **Preprocessing.**
   We assume the data is pre-processed to lie in $\mathbb{R}^D$. Local intrinsic dimensions are computed for each point as part of the pre-processing. With this, the data can be partitioned into sets of different intrinsic dimensions and steps 2, 3 and 4 can be applied to each partition separately.

Step 2 **Hierarchical Multi-scale Partitioning**
   We construct a hierarchy of partitions of the data using dyadic trees. The hierarchical partitioning provides a multiscale view of the data where the scale index is the resolution parameter. A stopping condition determines the leaf sets of the hierarchical partition. In our implementation, the stopping condition ensures that the local intrinsic dimension is smaller than or equal to the pointwise intrinsic dimension. Algorithmic tools for this construction include CoverTree [10] (which gives tree-like $\varepsilon$-nets with dyadically decreasing $\varepsilon$) or dyadic partitions, see, e.g., [16].

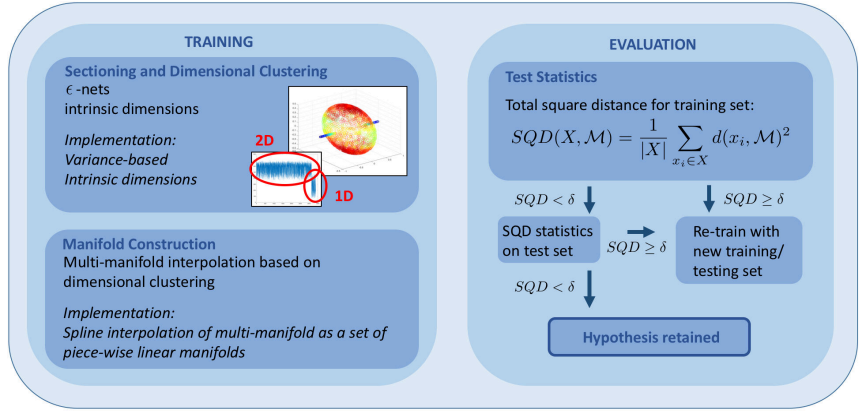

**Fig. 1:** Workflow for heuristic multi-manifold hypothesis test. We partition the data set using intrinsic dimensions directly computed from the data. Based on this, we construct multi-manifolds consisting of piece-wise linear manifolds that fit the data. The set of candidate multi-manifolds is then used to conduct a hypothesis test on the goodness-of-fit with the sample data.

Step 3  **Manifold construction.**
We perform a spline interpolation on the leaf sets of the partition-tree that gives piece-wise linear candidate manifolds consistent with the computed intrinsic dimensions. Coordinates associated with these piece-wise linear manifold can be used to construct non-linear splines to achieve a better goodness-of-fit.

Step 4  **Test statistics.**
We compute approximate square distances (Eq. 3) for the candidate multi-manifold. The total square distance is used as decision parameter for the hypothesis test.

## 4 Implementation

We implemented the methodology by defining algorithms for three functions:

- **Pre-processing.** A local intrinsic dimension $d$ for each point of a data set $X \subset \mathbb{R}^D$.
- **Multi-Manifold construction.** A dyadic linear multi-manifold $\mathscr{V}(X)$ approximating a data set $X$.
- **Test statistics.** A test statistic $S$ which takes as input a set $X$ of data points and a dyadic linear multi-manifold $\mathscr{V}$ and outputs the the expected value of the sum of the squared distances of $X$ and $\mathscr{V}$,

$$S(X, \mathscr{V}) = \mathbb{E}\left[SQD(S,M)\right] . \tag{1}$$

The workflow then consists of the following steps:

1. **Preprocessing.** Subdivide the sample points $X$ into a training and a testing set ($X_{\text{train}}$ and $X_{\text{test}}$). Compute local intrinsic dimensions for each data point in $X_{\text{train}}$. Stratify $X_{\text{train}}$ into strata $S_k$ using the local intrinsic dimensions.
2. **Multi-Manifold construction.** For each strata $S_k$, construct a dyadic linear multi-manifold $\mathscr{V}(S_k)$ that approximates the strata.
3. **Test statistics.** For each strata $S_k$, construct a probability distribution by applying the test statistics to the testing points $X_{\text{test}}$ of the data set and the dyadic linear multi-manifold $\mathscr{V}(S_k)$ that approximates the complementary training set $X_{\text{train}}$.

For higher accuracy, test statistics are averaged over multiple runs.

This implementation allows for testing the goodness-of-fit of a candidate multi-manifold $\mathscr{V}$. We sample a subset $S$ from the candidate multi-manifold $\mathscr{V}$ and compute intrinsic dimensions for each point in $S$. Based on these intrinsic dimensions, we stratify $S$ into strata $S_k$. Then, we construct a dyadic linear multi-manifold $\mathscr{V}(S_k)$ for each strata. For each value $k$ of the intrinsic dimension, the expected value $\mathbb{E}\left[SQD(S_k, \mathscr{V}(S_k))\right]$ of the sum of squared distances is computed and compared with the empirical distribution. If $SQD(S_k, \mathscr{V}(S_k))$ lies outside of the specified confidence interval, the hypothesis is rejected. For greater accuracy, the hypothesis test can be repeated multiple times. It there is no strata in the data set of the same intrinsic dimension as a strata $S_k$, the hypothesis is rejected for that strata of the candidate multi-manifold.

**Parameters: neighborhood definition.** It is clear that the method by which we define neighborhoods of points are essential for local estimation, both in terms of complexity and global estimation issues: While considering a small neighborhood can create computational errors and non-representative values, looking at a large neighborhood might cause global estimation issues. Here, we consider two types of neighborhood constructions, (i) neighborhoods consisting of balls centered at a design point and (ii) neighborhoods of the nearest neighbors of a design point. The size of the neighborhoods is chosen experimentally, we do not yet have a principled way to determine them.

### 4.1 Variance-based Local Intrinsic Dimension

In the current implementation, we used a variance-based local intrinsic dimension $d_{VLID}$. We define $d_{VLID}$ in terms of a *variance-based intrinsic dimension* $d_{VID}$, which takes as input a finite data set $X \subset \mathbb{R}^D$, a variance-based threshold $t \in [0,1]$ and a cutoff parameter $c$. If there are too few points in $N$, i.e. $|N| \leq c$, then $d$ is undefined. Otherwise, its output is the smallest integer $i$ such that the sum of the first $i$ squared singular values of the centered data set $N - \mathbb{E}(N)$ accounts for at least $t$ proportion of the total variance. In this case, $d_{VLID}$ is the PCA-based intrinsic dimension defined in Section 2.5. Recall that the total variance of a centered matrix is the sum of the squares of its singular values.

$$\mathrm{d}_{\mathrm{VLID}}(N) = \min_{1 \leq i \leq n} s.t. \left\{ \sum_1^i \sigma_j^2 \geq t \cdot \sum_1^n \sigma_j^2 \right\} \tag{2}$$

The *variance-based intrinsic dimension* depends on the parameters $t$ and $c$, and a list $L$ that determines a set of neighborhoods $N_i$ of (design) points in $X$. For example, $L$ could be a list of radii $r_i$ for neighborhoods $B(p, r_i)$ of radius $r_i$ centered at a design point $p$. For the nearest-neighbor based construction, $L$ could be a list of neighborhoods $KNN(p, k)$ consisting of the $k$-nearest neighbors of design points $p$. The value of the variance-based local intrinsic dimension function at a point $p$ is then defined as the minimum over the neighborhoods $N_i$ of the variance-based local intrinsic dimension $\mathrm{d}_{\mathrm{loc}}(N_i)$ whose cardinality exceeds the cutoff $c$:

$$\mathrm{d}(p, N_i) = \min_{1 \leq i \leq n} |N_i| > c \{\mathrm{d}_{\mathrm{VLID}}(N_i)\} . \tag{3}$$

The novelty of $d_{VLID}$ is its multiscale exploitation of projection-based intrinsic dimension, combined with a notion of cutoff.

## 4.2 Nearest Neighbors-based methods: Local GMST

An alternate method for computing intrinsic dimension is based on the GMST method applied locally. Suppose that we have a data set $X = \{x_1, x_2, \ldots, x_N\}$, where the samples points are drawn from a bounded density supported on a compact, $d$-dimensional Riemannian sub-manifold. Assume that this condition holds locally for some $n$ larger than a certain value $n^*$. Our local GMST algorithm uses the following steps:

1. Consider a point $x_i$ and construct a neighborhood $\mathcal{N}_{n,i}$ of $x_i$ using either a ball centered at $x_i$ containing, say, $n$-samples points or the $n$-nearest neighbors of $x_i$, $n > n^*$.
2. For each $x_i$ and the constructed neighborhood $\mathcal{N}_{n,i}$ above, find the $k$-Nearest Neighbors of each point $x_i$ in $\mathcal{N}_{n,i}$, where $k < n$. These form the sub-neighborhood $\mathcal{N}_{n,k,i}$.
3. Compute the total edge length of the kNN graph for each $\mathcal{N}_{n,i}$:

$$L_{\gamma,k}(\mathcal{N}_{n,i}) := \sum_{i=1}^n \sum_{x_j \in \mathcal{N}_{n,k,i}} |x_j - x_i|^\gamma,$$

where the parameter $\gamma$ determines locality. An equivalent formula if balls are used.

4. Using the fact that with probability 1 [21],

$$L_{\gamma,k}(\mathcal{N}_{n,i}) = a\, n^{\frac{d_{i,n}-\gamma}{d_{i,n}}} + \varepsilon_n, \tag{4}$$

where $\varepsilon - n$ gets small as $n$ grows and $a$ is some positive constant, the intrinsic dimension $d_{i,n}$ at each $x_i$ is found by applying non-linear least-squares.

Compute the intrinsic dimension $d_{i,n}$ for multiple neighborhoods $\mathcal{N}_{n,i}$ about $x_i$. The final intrinsic dimension at $x_i$ is found by averaging over the number of neighborhoods.

## *4.3 Dyadic Linear Multi-manifolds*

Given a data set $X \subset \mathbb{R}^D$, we recursively construct a sequence of linear multi-manifolds approximating the data set by recursively constructing a tree of dyadic cubes, such that the cubes at each level of the tree are disjoint, their union contains $X$ and approximating $X \cap C$ by the best fitting linear space $L_C$ of dimension $d_v(X \cap C)$ containing $E(X \cap C)$. Here $E(X \cap C)$ is the average of all of the points in $X \cap C$ and $d_v(X \cap C)$ is the variance-based dimension of $X \cap C$. This linear space can be computed using Singular Value Decomposition. Dyadic cubes are translates of cubes consisting of points whose $i^{th}$ coordinates lie in a dyadic interval $[0, 2^{-k_i}]$. For the root of the tree, choose a cube which contains $X$. To obtain the other cubes, choose an order of the coordinate and sequentially divide the cube in half along a specific coordinate axis. Recursively cycle through the sequence of coordinates. This results in a binary tree, making the computation easier, although a tree can also be constructed by halving all of the sides of the parent cube (not just one side).

The depth of the tree varies with the stopping condition used in the algorithm. Different stopping conditions for the recursive algorithm determine different dyadic linear multi-manifolds $\mathcal{V}(X)$. In our implementation we constructed the dyadic linear multi-manifolds for subsets $X(i) \subset X$ that consists of all points with local intrinsic dimension $i$. In this case, we could use the the stopping condition that the variance-based intrinsic dimension of the leaves is smaller than $i$. The sets $L_C \cap C$, corresponding to the leaf cubes form the candidate multi-manifold $\mathcal{V}(X(i))$. We defined $\mathcal{V}(X)$ as the union of the dyadic linear multi-manifolds $\mathcal{V}(X(i))$ for each intrinsic dimension.

## 4.4 Estimating the Sum of Squared Distances function SQD

In the current implementation we exploit the variance-based definition of the intrinsic dimension ($d_v$). We observe that for any data set $X$ the squared distances to its best linear space $L$ of dimension $d_v$ is bounded above by the sum of singular values:

$$SQD(X,L) < (1-t) \sum_{i>d_v} \sigma_i^2 . \qquad (5)$$

We used this observation to define $SQD(S,\mathcal{V})$: We define $SQD$ for each multi-manifold component, i.e. for each linear space $L_C$ in the dyadic linear multi-manifold.

$$SQD_{(L_C)} = (1-t) \sum_{i>d_{\text{loc}}(S \cap A)} \sigma_i^2 . \qquad (6)$$

In this equation, $\sigma_i$ is the $i^{th}$ singular value of the centered data set, $S \cap L_C - \mathbb{E}(S \cap L_C)$. Then the sum of squared distances function from a data set $S$ to a multi-manifold $\mathcal{V}$ consisting of components $L_C$ is defined in the current implementation by summing up the sum of squared distances functions for each of the components:

$$SQD(S,\mathcal{V}) = \sum_{(L_C) \sim \mathcal{V}} SQD_{(L_C)} . \qquad (7)$$

## 5 Experimental Validation

We demonstrate the methodology for two low-dimensional use cases. For both cases, we compute intrinsic dimension, construct a candidate multi-manifold and compute test statistics. The first data set consists of a simple simple sphere-line object with components of different intrinsic dimensions: A one-dimensional line, a two-dimensional surface and three dimensional intersection points (see Figure 2).

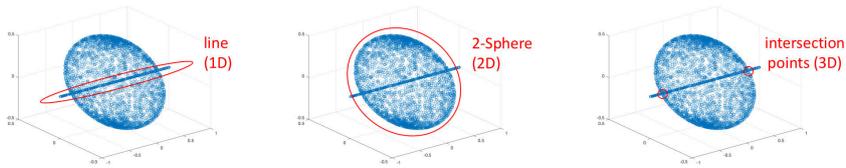

**Fig. 2**

The second data set consists of three-dimensional coordinates for a LiDAR image of the Golden Gate Bridge (see Figure 8). Intuitively, the bridge cables appear to be 1-dimensional, while the surface of the bridge should to be 2-dimensional. We will test this intuition in the following analysis.

## *5.1 Use Case: Sphere - Line*

The data set consists of a sample from a Sphere-Line configuration (see Fig. 2). For ease of computation, only points on the sphere and the line segments external to the sphere were sampled. We first computed the intrinsic dimension of the sample points. As shown in Fig. 2, we would expect to find points of intrinsic dimension 1 (line) and 2 (sphere surface) and two points of intrinsic dimension 3 (intersection points). The sphere is curved, so samples from its surface will not be well-approximated by a linear multi-manifold. We sampled randomly using polar coordinates on the sphere in order to preserve the intrinsic geometry to the best possible extend. The sample $X$ consisted of 2708 points; 2513 from the sphere and 193 from the line external to the sphere. The sampled sphere is of radius $\frac{1}{2}$ and centered at the origin; the line sample was randomly selected from the intervals [-1,-1/2] and [1/2,1] on the x axis. Because polar coordinates were used, the sampling from the sphere was not uniform with respect to the surface area measure. First, the intrinsic dimension of the points in $X$ was computed using the variance-based intrinsic dimension algorithm discussed in Section 4.1 with neighborhoods of radii 2 to 0.1 in decrements of 0.1. The intrinsic-dimension based strata have the following cardinalities: $|X(1)| = 157$, $|X(2)| = 2514$ and $|X(3)| = 37$. The sample points are shown in Fig. 3, color coded by intrinsic dimension values.

For each of the intrinsic-dimension based strata $X(i)$, a dyadic linear multi-manifold $\mathcal{V}(X(i))$ is computed approximating the strata. A summary of the properties of each of the multi-manifolds is shown in Table 1.

The multi-manifold for the points with intrinsic dimension one ($\mathcal{V}(X(1))$) has only one linear component, which agrees with the fact that all of these points are on the x-axis. In this sample, the multi-manifold for the intrinsic dimension two points $\mathcal{V}(X(2))$ has is two linear components, one corresponding to the cube consisting of points $x \leq 0$ and the other cube consisting of points $x > 0$. The somewhat surprising fact is that the local intrinsic dimension of the sphere samples in each of these halves of the unit cube is one. In this example the parameter is $t = 0.95$. The multi-manifold $\mathcal{V}(X(3))$ for the 37 points of intrinsic dimension three also had only one component. This is explained by Figure 3 which shows that most of the points of intrinsic dimension three are on the x-axis, a 1-dimensional linear manifold, near the points of intersection with the sphere. To summarize the goodness-of-fit of the linear multi-manifolds, we compute the expected value of the squared distances of the data points in the cube to the best fitting linear affine space of the local intrinsic dimension $d$.

The last step of the methodology is the computation of a probability distribution $H(i)$ for each value of the intrinsic dimension $i$. For the sphere-line example, this was done by randomly choosing 20 test subsets for each i, which determined

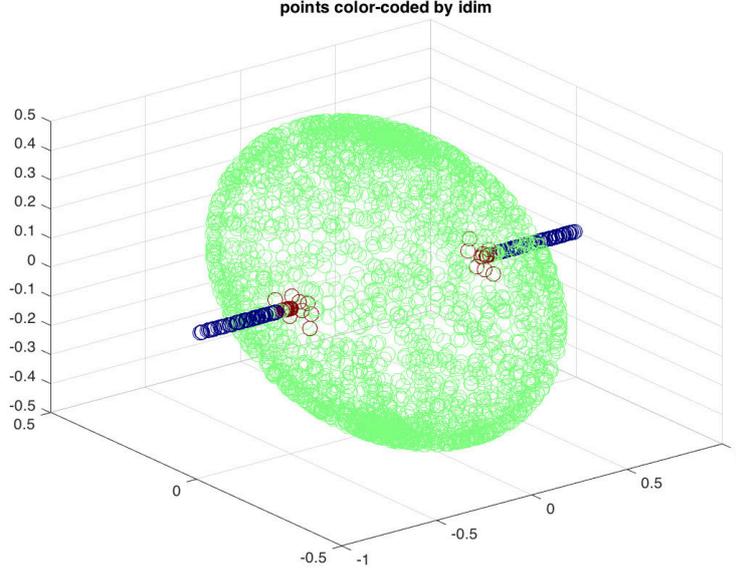

**Fig. 3:** Intrinsic Dimensions of the Sphere-Line Sample.

| dim | total pts | $\mathscr{V}$ points | components | $\mathbb{E}(SQD)$ |
|---|---|---|---|---|
| 1 | 157 | 157 | 1 | 0.0076 |
| 2 | 2514 | 2514 | 2 | 0.0096 |
| 3 | 37 | 37 | 1 | 0.0031 |

**Table 1:** Summary of the Multi-Manifolds $\mathscr{V}(X(i))$ for the Sphere-Line Example. The intrinsic dimension is shown in the first column, the number of points in each strata in the second. The third column shows the total number of points; this is the sum of the points that lie in the dyadic cubes associated with each component of the multi-linear manifold. In this case, the total number of points equals the number of supported points for each strata since the sampling was fairly dense. However, that will not be true in general since in the top-down recursive algorithm no component of the multi-manifold will be constructed if there are less than $Klog(K)$ points. Here K was specified to be three, since that was the maximum intrinsic dimension. The fourth column lists the number of components of $\mathscr{V}(X(i))$.

20 training subsets $X_{train}(i) = X(i) - X_{test}(i)$. The dyadic linear multi-manifold $\mathscr{V}(X(i))$ was computed for each training set and the expected value $\mathbb{E}_i(SQD)$ of the sum of the squares of the distances of the test subset $X_{test}(i)$ to $\mathscr{V}(X(i))$ was computed (cube by cube). The hypothesis testing probability distribution $H(i)$ is the distribution of the statistics $\mathbb{E}_i(SQD)$. The expected value, standard deviation, and z-score cutoff for a confidence interval were computed for a 95% confidence interval. The computed values for these statistics are shown in Tab. 2. The informa-tion in Tab. 2 is sufficient to make a hypothesis testing decision for each intrinsic dimension.

| d | $\mathbb{E}(\mathbb{E}(SQD))$ | support | train count | test count | runs | $SD(\mathbb{E}(SQD))$ | z cutoff |
|---|---|---|---|---|---|---|---|
| 1 | 0 | 1 | 1814 | 893 | 20 | 0 | 0.3110 |
| 2 | 0.0541 | 1 | 1814 | 893 | 20 | 0.0007 | 1.5190 |
| 3 | 0.0011 | 1 | 1814 | 893 | 20 | 0.0004 | 1.7146 |

**Table 2:** Test Statistics for the Sphere-Line Sample. The third column shows the expected value of the number of points in the dyadic cubes supporting the multi-manifold. The fourth and fifth column show the expected values of the sizes of the training and testing sets.

We also implemented a simple version of local GMST. We use the first algorithm (using Laurens van der Maaten's implementation) and compute the intrinsic dimension of each point in the data sets using neighborhood $\mathscr{N}_{n,i}$ of size $n$ in the range 200 to 400, with increments of 25. We only performed the experiment on the sphere-line data set as the computation time grows prohibitively large while the results obtained do not match all our predictions, see Figure (14). Our results show that the points on the lines are dimension 1 and the points around the intersection of the line and the sphere have dimension around 3. However, this is the case for several points on the other parts of the balls as well. The results are thus poorer than those obtained with the Variance Based estimator. This is because for the GMST, the size $n$ of $\mathscr{N}_{n,i}$ for each $x_i$ has to be large enough for the guarantee 4 to hold. It is clear that the method used to construct the neighborhood of a point is essential for local estimation, both from the point of view of complexity but also, because picking a small neighborhood can create computational errors and non-representative values while for large neighborhood, the method will carry global estimation issues. At this stage, we do not have a principled way to find these sizes.

In future work, we could vary we also hope to vary the $k$ and $n$ parameters for the GMST, but also $\gamma$. We also hope to implement the MSVD algorithm [15, 35].

## 5.2 Use Case: LiDAR data

To provide the context for our LiDAR data use case, we summarize LiDAR technology, example applications, the LiDAR data collection process and the measurements taken in the process. We then describe the specific use case in this context.

LiDAR stands for light detection and ranging and it is an optical remote sensing technique that uses laser light to densely sample the surface of the earth, producing highly accurate $x$, $y$ and $z$ measurements. The resulting mass point cloud data sets can be managed, visualized, analyzed and shared using ArcGIS. The collection vehicle of LiDAR data might be an aircraft, helicopter, vehicle and tripod.

LiDAR is an active optical sensor that transmits laser beams towards a target while moving through specific survey routes. The reflection of the laser from the target is detected and analyzed by receivers in the LiDAR sensor. These receivers record the precise time from when the laser pulse leaving the system to when it returns to calculate the range distance between the sensor and the target, combined with the positional information GPS (Global Positioning System), and INS (inertial navigation system). These distance measurements are transformed to measurements of actual three-dimensional points of the reflective target in object space.

LiDAR can be applied, for instance, to update digital elevation models, glacial monitoring, detecting faults and measuring uplift detecting, forest inventory, detect shoreline and beach volume changes, landslide risk analysis, habitat mapping and urban development [36]. 3D LiDAR point clouds have many applications in the Geosciences. A very important application is the classification of the 3D cloud into elementary classes. For example, it can be used to differentiate between vegetation, man-made structures and water. Alternatively, only two classes such as ground and non-ground could be used. Another useful classification is based on the heterogeneity of surfaces. For instance, we might be interested classifying the point cloud of reservoir into classes such as gravel, sand and rock. The design of algorithms for classification of this data using a multi-scale intrinsic dimensionality approach is of great interest to different scientific communities [11] [3].

The LiDAR data considered here was converted to 3D coordinates, using the free QGIS software. It contains approximately 87,000 points, a scatter plot is shown in Figure 8. In terms of dimensionality, the catenary cables at the top of the bridge should have intrinsic dimension one and the bridge surface intrinsic dimension two. We will test this intuition using the multi-manifold testing framework.

The point data is post-processed after the LiDAR data collection survey into highly accurate geo-referenced $x$, $y$, $z$ coordinates by analyzing the laser time range, laser scan angle, GPS position, and INS information. We have followed very closely the exposition in [46] and [48].

**LiDAR point attributes** The following attributes along with the position $(x,y,z)$ are maintained for each recorded laser pulse. We have included a description of each attribute and complemented the intensity attribute description with the exposition in [46][2].

- Intensity. Captured by the LiDAR sensors is the intensity of each return. The intensity value is a measure of the return signal strength. It measures the peak amplitude of return pulses as they are reflected back from the target to the detector of the LiDAR system.
- Return number. An emitted laser pulse can have up to five returns depending on the features it is reflected from and the capabilities of the laser scanner used

---

[2] The description of each of the attributes below are literally taken from website `http://desktop.arcgis.com/en/arcmap/` (in 'Fundamentals about LiDAR under 'Manage Data')

to collect the data. The first return will be flagged as return number one, the second as return number two, and so on.

- <u>Number of returns</u>. The number of returns is the total number of returns for a given pulse. Laser pulses emitted from a LiDAR system reflect from objects both on and above the ground surface: vegetation, buildings, bridges, and so on. One emitted laser pulse can return to the LiDAR sensor as one or many returns. Any emitted laser pulse that encounters multiple reflection surfaces as it travels toward the ground is split into as many returns as there are reflective surfaces.
- <u>Point classification.</u> Every LiDAR point that is post-processed can have a classification that defines the type of object that has reflected the laser pulse. LiDAR points can be classified into a number of categories including bare earth or ground, top of canopy, and water. The different classes are defined using numeric integer codes in the LAS files.
- <u>Edge of flight line</u>. The points will be symbolized based on a value of 0 or 1. Points flagged at the edge of the flight line will be given a value of 1, and all other points will be given a value of 0.
- <u>RGB</u>. LiDAR data can be attributed with RGB (red, green, and blue) bands. This attribution often comes from imagery collected at the same time as the LiDAR survey.
- <u>GPS time</u>. The GPS time stamp at which the laser point was emitted from the aircraft. The time is in GPS seconds of the week.
- <u>Scan angle.</u> The scan angle is a value in degrees between -90 and +90. At 0 degrees, the laser pulse is directly below the aircraft at nadir. At -90 degrees, the laser pulse is to the left side of the aircraft, while at +90, the laser pulse is to the right side of the aircraft in the direction of flight. Most LiDAR systems are currently less than $\pm 30$ degrees.
- <u>Scan direction.</u> The scan direction is the direction the laser scanning mirror was traveling at the time of the output laser pulse. A value of 1 is a positive scan direction, and a value of 0 is a negative scan direction. A positive value indicates the scanner is moving from the left side to the right side of the in-track flight direction, and a negative value is the opposite.

Points clouds are are a very dense collection of points over an area. A laser pulse can be returned many times to the airborne sensor. See figure 4 for graphic explanation of this process with a tree.

In the case of a simple laser profiler that has been mounted on an airborne platform, the laser points vertically toward the ground to allow a rapid series of measurements of the distances to the ground from the successive positions of the moving platform. The measurements of the vertical distances from the platform to a series of adjacent points along the ground track are made possible through the forward motion of the airborne or space-borne platform. If the positions and altitudes of the platform at these successive positions in the air or in space are known or can be determined (e.g., using a GPS/IMU system), then the corresponding ranges measured at these points will allow their ground elevation values to be determined. Consequently, these allow the terrain profile along the flight line to be constructed (see figures 5 and 6).

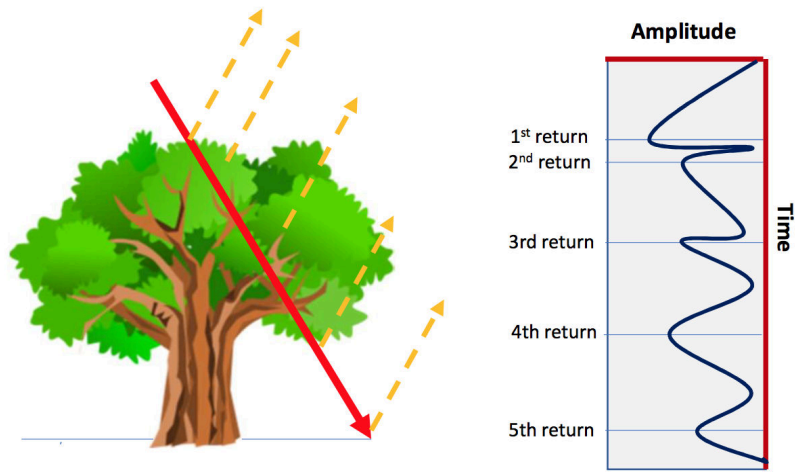

**Fig. 4:** A pulse can be reflected off a tree's trunk, branches and foliage as well as reflected off the ground. The image is recreated from a figure in [48], pp. 7

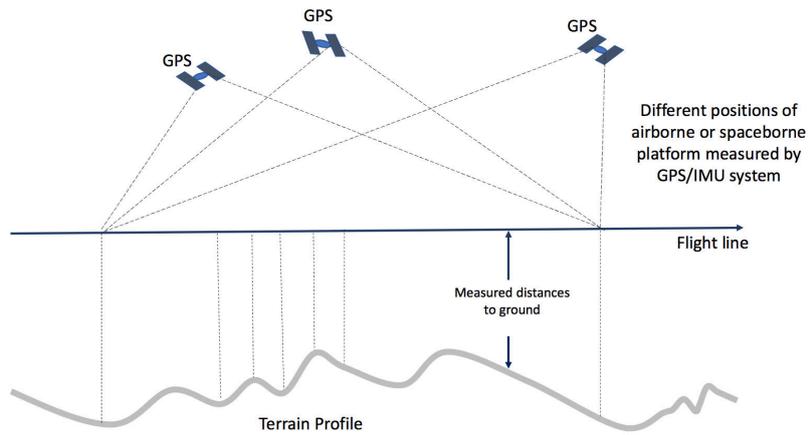

**Fig. 5:** Profile being measured along a line on the terrain from an airborne or space-borne platform using a laser altimeter. The image reproduced from [46], Chapter 1, pp. 7

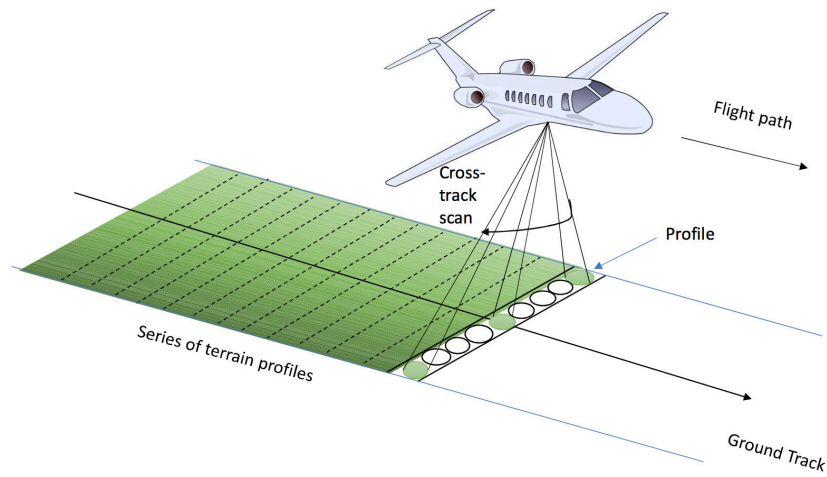

**Fig. 6:** The profile belonging to a series of terrain profiles is measured in the cross track direction of an airborne platform. The image was recreated from figure 1.5 (b), pp. 8 in [46].

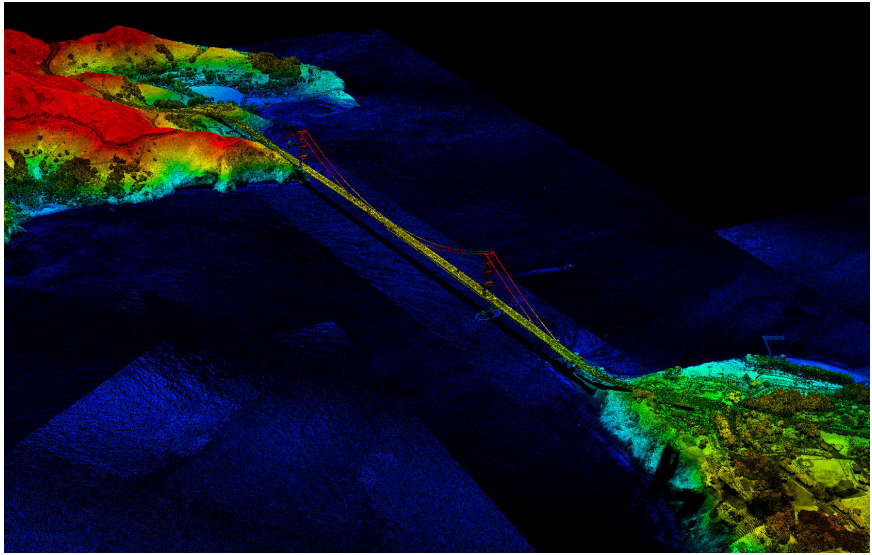

**Fig. 7:** 3D point cloud LiDAR visualization of the Golden Gate Bridge, San Francisco, CA. The image was produced by Jason Stoker (USGS) using LP360 by Qcoherent [47].

For our use case, we use LiDAR data from the Golden Gate Bridge, San Francisco, CA. We extracted the original data (more than eight million points) from the *USGS EarthExplorer* (https://earthexplorer.usgs.gov/) and sampled using the software QGIS. Fig. 7 illustrates a visualization of the 3D point cloud data of the complete bridge. We just worked with one part of the bridge and the surrounding ground, vegetation and water (see Fig. 8). We did not work with all the above mentioned attributes, but extracted only spacial coordinates $x$, $y$, $z$ for our study.

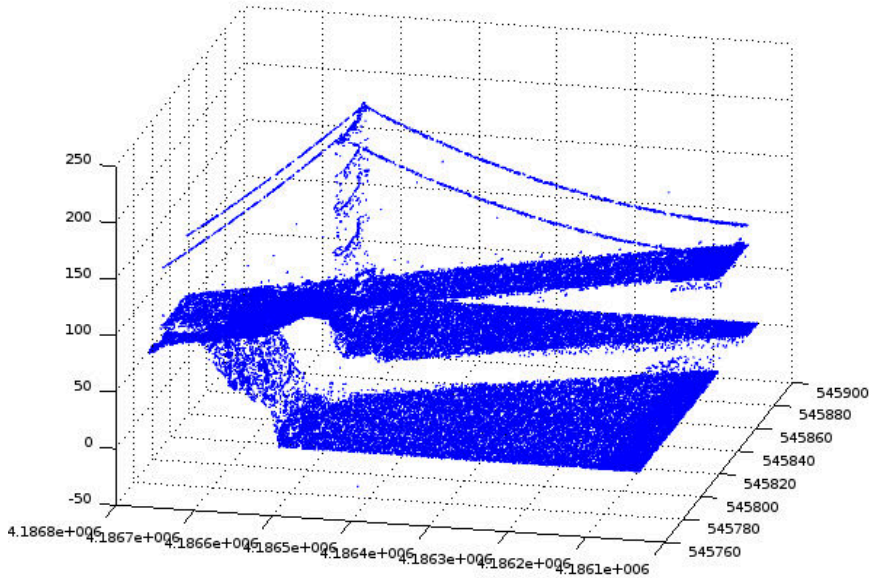

**Fig. 8:** Scatter plot of the Golden Gate Bridge section of the data. We extracted $\sim 87,000$ points from the original LAS file for the analysis.

### 5.2.1 Intrinsic Dimension Results

The data was pre-processed by computing the variance-based local intrinsic dimension using balls of dyadic scales 4 through 7. Specifically, we used neighborhoods of radii $\text{diam} \cdot 2^{-\text{scale}}$ for scale $= 4...7$, where the diameter was the maximum of the coordinate diameters. Figure 9 shows that the catenary cables indeed have intrinsic dimension one, the surface of the bridge has intrinsic dimension two, and the intersection of the main catenary cables with the bridge columns have dimension three. This confirms our intuition on the intrinsic dimension.

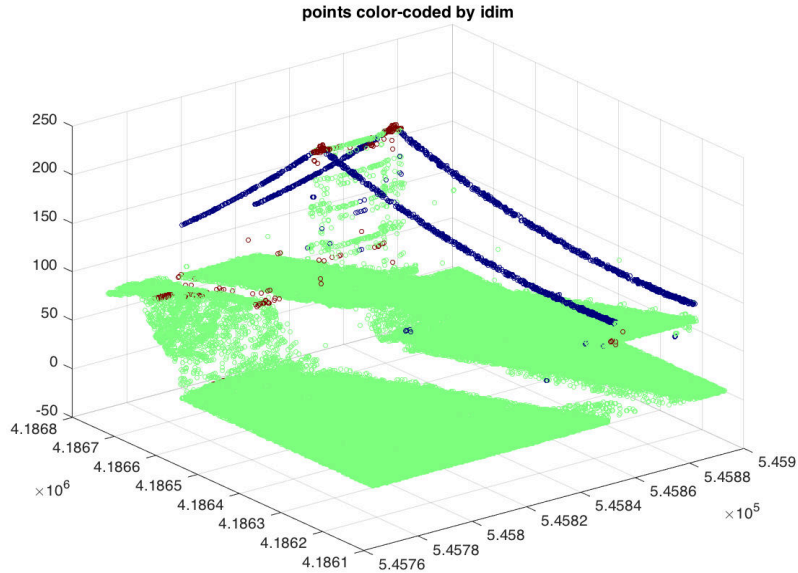

**Fig. 9:** LiDAR data set of the Golden Gate Bridge.

Two additional views were computed and visualized using variance-based local intrinsic dimension based color-coding for further understanding the data set. In Fig. 10 the data set was colored by a lexicographic ordering of the intrinsic dimension, minus the ordering of the radii. In Fig. 11 the data was colored by the expected value of the total variance over the radii at which the intrinsic dimension was observed. Figures 10 and 11 show more subtle distinctions than are revealed by the intrinsic dimension statistics, but these were not used in the remainder of the analysis. This experiment demonstrates that meaningful geometric structures can be inferred from analyzing intrinsic dimensions in densely sampled low-dimensional data.

We also computed the intrinsic dimension using the Variance-Based estimator for the same data sets, but this time, we formed the neighborhoods using the $k$ nearest neighbors of a given point $x$. Our results are practically identical to the ones obtained when the neighborhoods are formed using balls of radius $r$ centered at $x$. The main advantage of $k$ nearest neighbors is that we are guaranteed that the neighborhoods considered for the intrinsic dimension estimation are not empty. For the sphere-line example in Figure (12), the entire sample size consisted in 2708 points and we computed the intrinsic dimension for neighborhood size in the range 50 to 700, with increments of 25. For the LiDAR data in Figure (**??**), we used a range of neighborhood sizes of 50 to 800, with increments of 50. The same conclusion as for the sphere-line holds, i.e., for most points, the intrinsic dimension obtained for each point are identical as those obtained using balls.

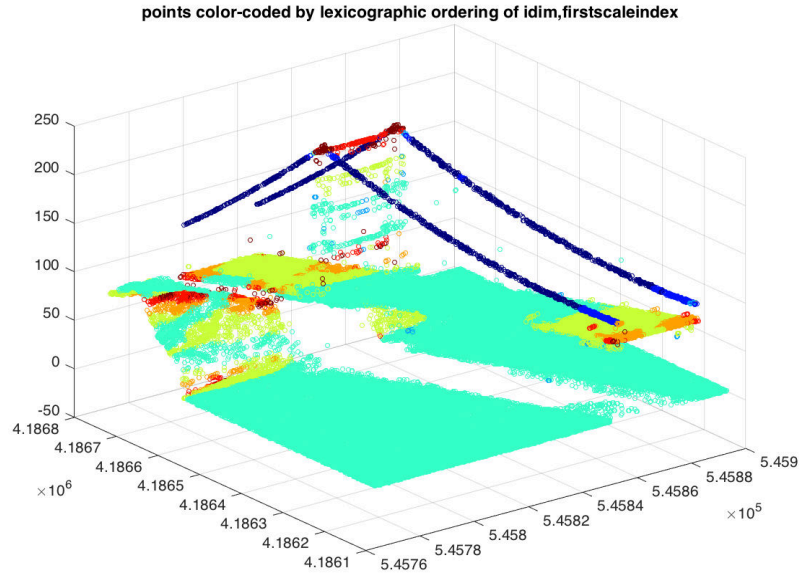

**Fig. 10:** LiDAR data set of the Golden Gate Bridge

### 5.2.2 Hypothesis Construction and Testing Results for the LiDAR Data Set

For each of the three intrinsic-dimension based strata $D(i)$ of the LiDAR data, a dyadic linear multi-manifold $\mathscr{V}(D(i))$ was computed approximating the strata. A summary of the properties of each of the multi-manifolds is shown in Table 3. There is one row for each intrinsic dimension.

| dim | totalpts | MMpoints | components | EVsqdist |
|---|---|---|---|---|
| 1 | 1891 | 1885 | 20 | 0.0003 |
| 2 | 84698 | 84698 | 66 | 0.0002 |
| 3 | 1185 | 1185 | 1 | 0.0079 |

**Table 3:** Summary of the Multi-Manifolds $\mathscr{V}(D(i))$ for the LiDAR data.

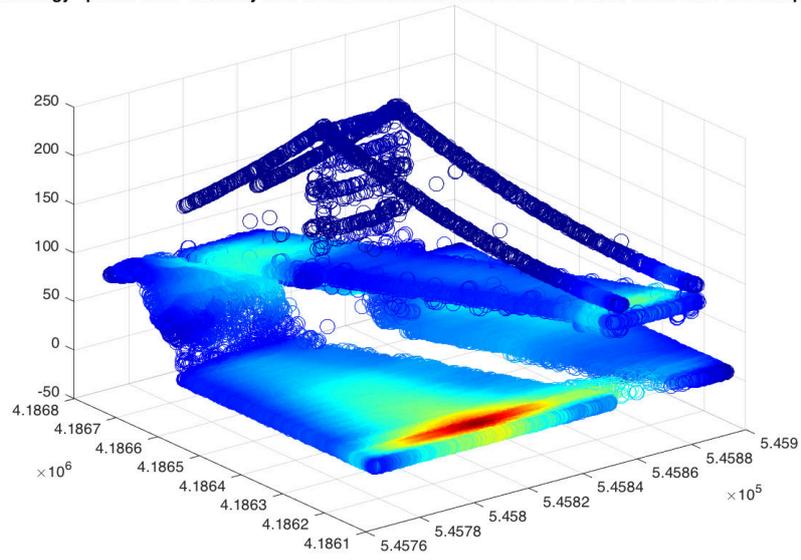

**Fig. 11:** LiDAR data set of the Golden Gate Bridge

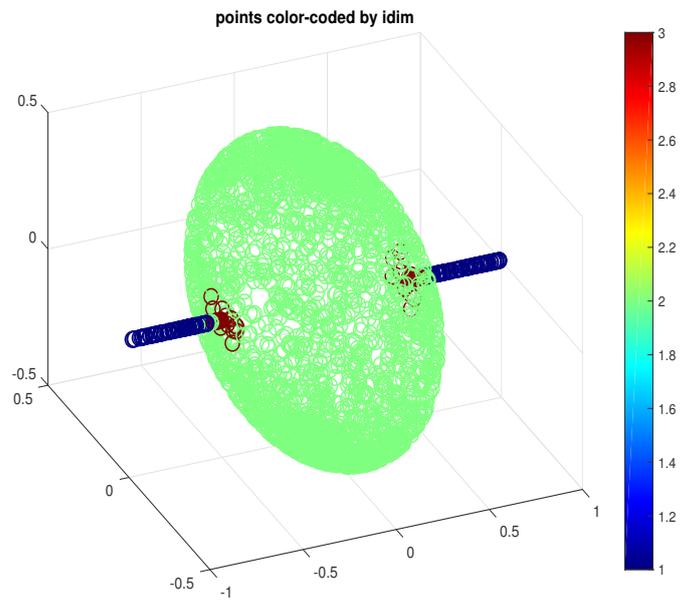

**Fig. 12:** Variance-Based estimator with k-nearest neighbors for sphere-line

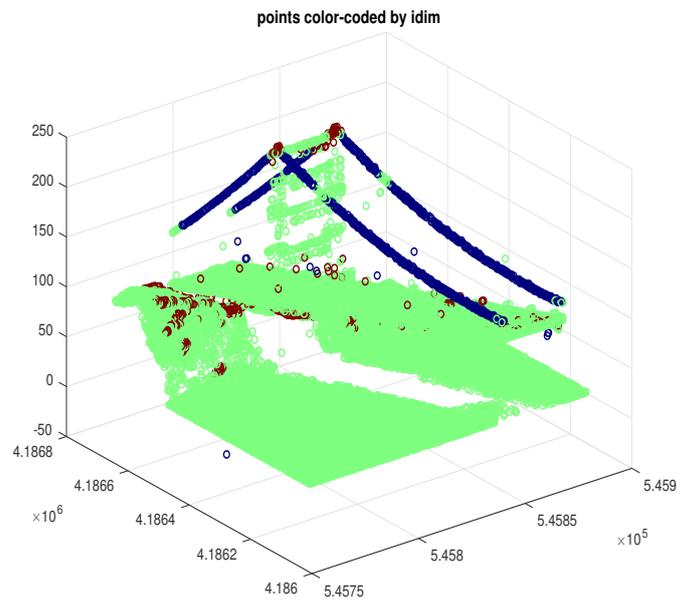

**Fig. 13:** Variance-Based estimator with k-nearest neighbors for LiDAR data

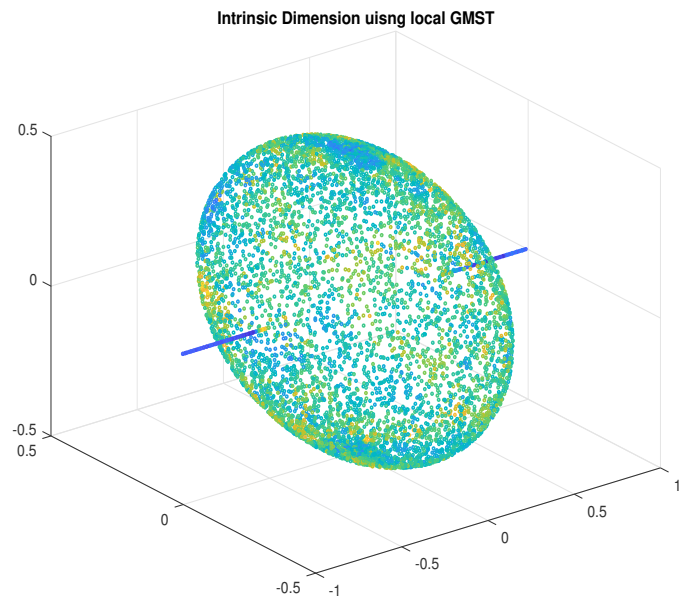

**Fig. 14:** LOCAL GMST k-nearest neighbors for sphere-line

Finally, we computed of a probability distribution $H(i)$ for each intrinsic dimension value. As for the Sphere-Line example, this was done by randomly sampling testing and training subsets. The results are shown in Table 4.

| d | $\mathbb{E}(\mathbb{E}(SQD))$ | support | train count | test count | runs | $SD(\mathbb{E}(SQD))$ | z cutoff |
|---|---|---|---|---|---|---|---|
| 1 | 2.9926 | 1 | 58834 | 28940 | 20 | 0.17154 | 1.4401 |
| 2 | 3.9316 | 1 | 58834 | 28940 | 20 | 0.21638 | 1.7247 |
| 3 | 0 | 1 | 58834 | 28940 | 20 | 0 | 1.5343 |

**Table 4:** Test Statistics for the LiDAR data.

# 6 Future Research: Questions and Directions

This article presents a summary of conceptual ideas and preliminary results from a workshop collaboration. In line with the exploratory style of the article, we outline a number of further research questions and possible future directions:

1. For what data sets and applications is multi-manifold hypothesis testing useful in practice? The examples in this paper are limited to densely sampled low-dimensional data sets. – How does the method perform on higher dimensional data sets and on sparse data sets (e.g. Word2Vec)?
2. Could intrinsic dimension statistics be used to find change points or change boundaries (commonly used in statistics)? Can the dyadic linear multi-manifold structure be useful for the formulation of high-dimensional trends for multi-dimensional time series and high-dimensional change boundary detection?
3. Can a dyadic linear multi-manifold structure be exploited to construct a non-linear multi-manifold which models the data more accurately, has known smoothness properties, and has as few components as possible?
4. What are the most practical and effective methods for improving the scalability of the intrinsic dimension computation? What additional state-of-the art algorithms can be exploited to realize computationally efficient hypothesis testing for multi-manifolds?
5. How robust is the presented approach? The investigations could include robustness to changes in the tree structure, the neighborhood choices and changes in the intrinsic dimension algorithm itself.
6. Are there additional or alternative test statistics which could be efficiently computed to compare samples of candidate multi-manifolds and the constructed training manifolds, for example test statistics that compare structural properties?
7. Could computational topology be used to estimate the optimal number of manifold components and the minimal number of patches?

8. Are there conditions on a data set as a subset of a tree-structured space which will guarantee that the total variance for node subsets associated with a level in the tree is monotonically decreasing as the distance of the level from the root increases?
9. How could the theory of quantitative rectifiability be exploited or enhanced to provide theoretical guarantees for multi-manifold hypothesis testing?

## 7 Summary and Conclusions

In this paper we present conceptual ideas and preliminary results for the development of a heuristic framework for multi-scale testing of the multi-manifold hypothesis:

Given a data set $X = \{x_i\}_{i \in I}$ in $\mathbb{R}^n$ and a multi-manifold $\mathscr{V}$, is the square distance of the points in $X$ to the multi-manifold $\mathscr{V}$ more than one would expect? If so, we reject $\mathscr{V}$ as being a multi-manifold that fits $X$. We describe an implementation of this heuristic framework and demonstrate it on two low-dimensional, densely sampled data sets with intuitive geometry. The experiments demonstrate that the computed low-dimensional multi-manifold is consist with the intuitive geometry.

Our approach exploits fundamental methods of statistical reasoning, hypothesis testing and simple variance-based analysis, as well as multi-scale representation methods. We apply summary statistics to data computed at multiple scales using results from geometric representation theory. The specific distribution is computed empirically from the data.

We expect that many other algorithms can be exploited in alternative realizations of the framework. Further directions that could build on our approach are outlined at the end of the paper. To ensure the reproducibility of our results, the prototype implementation will be made publicly available on GitHub.

**Acknowledgements** This research started at the Women in Data Science and Mathematics Research Collaboration Workshop (WiSDM), July 17-21, 2017, at the Institute for Computational and Experimental Research in Mathematics (ICERM). The workshop was partially supported by grant number NSF-HRD 1500481-AWM ADVANCE and co-sponsored by Brown's Data Science Initiative.

Additional support for some participant travel was be provided by DIMACS in association with its Special Focus on Information Sharing and Dynamic Data Analysis. LN worked on this project during a visit to DIMACS, partially supported by the National Science Foundation under grant number CCF-1445755 and by DARPA SocialSim-W911NF-17-C-0098. FPM received partial travel funding from Worcester Polytechnic Institute, Mathematical Science Department.

We thank Brie Finegold and Katherine M. Kinnaird for their participation in the workshop and in early stage experiments. In addition, we thank Anna Little for helpful discussions on intrinsic dimensions and Jason Stoker for sharing material on LiDAR data.

## Code Availability

An implementation of the workflow in MATLAB is available on GitHub:
`https://github.com/MelWe/mm-hypothesis`.